%% file: main.tex
\documentclass[runningheads]{llncs}

\usepackage{comment,authblk,graphicx}
\usepackage{subcaption}
\usepackage{algorithm}
\usepackage{amsmath}

\usepackage{algpseudocode}
\usepackage{graphicx}
\usepackage{array}
\usepackage{caption} 
\usepackage{makecell}

\begin{document}

\title{Regularized Contrastive Learning of Semantic Search}
\author{Mingxi Tan\thanks{Corresponding author} \and Alexis Rolland \and Andong Tian}
\authorrunning{M. Tan et al.}
\institute{Ubisoft La Forge, Ubisoft, China\\
\email{\{ming-xi.tan,alexis.rolland,an-dong.tian\}@ubisoft.com}}

\date{}

\maketitle

\input{content}


\end{document}

%% file: content.tex
\begin{abstract}
Semantic search is an important task which objective is to find the relevant index from a database for query. It requires a retrieval model that can properly learn the semantics of sentences. Transformer-based models are widely used as retrieval models due to their excellent ability to learn semantic representations. in the meantime, many regularization methods suitable for them have also been proposed. In this paper, we propose a new regularization method: Regularized Contrastive Learning, which can help transformer-based models to learn a better representation of sentences. It firstly augments several different semantic representations for every sentence, then take them into the contrastive objective as regulators. These contrastive regulators can overcome overfitting issues and alleviate the anisotropic problem. We firstly evaluate our approach on 7 semantic search benchmarks with the outperforming pre-trained model SRoBERTA. The results show that our method is more effective for learning a superior sentence representation. Then we evaluate our approach on 2 challenging FAQ datasets, Cough and Faqir, which have long query and index. The results of our experiments demonstrate that our method outperforms baseline methods.
\keywords{contrastive \and regularization \and semantic.}
\end{abstract}

\vspace{-2.5em}
\section{Introduction}
On a semantic search task, we generally have a dataset which consists of tuples of sentences $<$Query, Index$>$. The objective of the task is to find the best indices for each input query on the dataset. First, we need to calculate the embeddings of the input queries and the candidate index, then we compute their cosine similarity scores, and finally rank the candidate answers based on these scores. Therefore, superior embeddings are crucial to the task.

Recently, transformer-based models are widely used to perform this kind of task due to their powerful semantic representation capabilities. They use contextual information to create embeddings that are sensitive to the surrounding context~\cite{Arora}. However, learning a better sentence embedding is still a challenging task. Many regularization methods suitable for Transformers have been proposed such as attention dropout and DropHead~\cite{Wangchunshu}. They all can  reduce overfitting and improve performance. In this work, we introduce a new regularization method: Regularized Contrastive Learning (RCL), which augments the number of semantically similar embeddings of each sentence and uses them to construct contrastive regulators for learning better sentences semantics.

\begin{figure}[htp]
\centering
\begin{subfigure}{\textwidth}
\centering
\includegraphics[width=7cm]{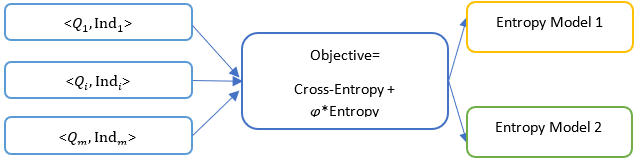}

\vspace{-1em}
\caption{Train entropy model}
\label{fig:f1a}
\end{subfigure}
\begin{subfigure}{\textwidth}
\centering
\includegraphics[width=7cm]{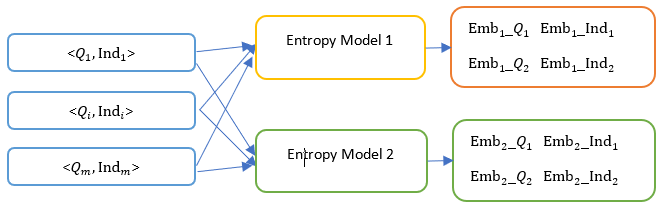}
\vspace{-1em}
\caption{Generate semantically similar embedding of each sentence through entropy model}
\label{fig:f1b}
\end{subfigure}
\begin{subfigure}{\textwidth}
\centering
\includegraphics[width=7cm]{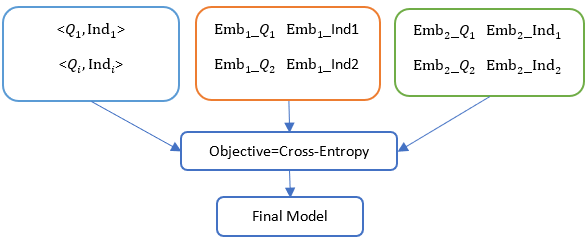}

\vspace{-0.5em}
\caption{Take the embeddings of tuples in the database and their corresponding augmented embeddings into the contrastive objective to train the final model}
\label{fig:f1c}
\end{subfigure}
\caption{Regularized Contrastive Learning.}
\label{fig:f1}
\vspace{-2em}
\end{figure}

RCL mainly contains 2 steps as shown in Figure \ref{fig:f1}. In the first step, we train several models by adding an entropy term to the contrastive objective, denoted as entropy model (Figure \ref{fig:f1a}). 
Each of the entropy model can generate a semantically similar embedding for each sentence (Figure \ref{fig:f1b}). The role of this step is to augmente the data. Different from the other widely used methods such as sentence cropping, word replacement and word deletion, which require extra effort to ensure the altered sentences have similar semantics to the original sentences, we do not directly augment the number of sentences but the number of embeddings. Unlike LAMBADA (language-model-based data augmentation) \cite{Ateret} which augments the number of embeddings based on GPT-2 according to specific class, our method augments the number of semantically similar embeddings. The second step is to take the embeddings of tuples $<$Query, Index$>$ in the dataset and their generated embeddings into contrastive objective to fine-tune the final model (Figure 
\ref{fig:f1c}).
\vspace{-2em}
where  \(<Q_i, Ind_i>\) represents a query and index sentence tuple, \(Emb_k\_Q_i\) represents the embedding of query i computed by entropy model k, \(Emb_k\_Ind_i\) represents the embedding of index i computed by entropy model k. \(\phi\) is the weight of entropy term.

We evaluate our approach on seven semantic search benchmarks and two challenging information retrieval tasks. Information retrieval (IR) task is the direct application of semantic search. Using a FAQ dataset, which consists of several tuples $<$Question, Answer$>$, our objective is to find the best answers for each input question on the given dataset.
The results of our experiments show that our method is effective of learning superior sentences embeddings compared to other regularization methods such as attention dropout and DropHead.
The main contributions of this paper are summarized as follows:

\begin{enumerate}
  \item A Regularized Contrastive Learning method that augments the number of sentence embeddings and takes them into the contrastive objective as regulator.
  \item Experiments results on 7 standard semantic textual similarity tasks demonstrate our method is efficient for semantic comprehension.
  \item Experiments results on 2 IR tasks show that our method can improve the performance of IR models. 
\end{enumerate}

\section{Related Work}
\textbf{Semantic search tasks} require models with excellent semantic understanding, which is different from translation and text-classification tasks. They use the contrastive objective \cite{Su,Gao,Reimers}, shown in Eq.\ref{eq1}, because the corresponding dataset directly provides semantically similar tuples of $<$Query, Index$>$. 
\begin{equation} \label{eq1}
\resizebox{0.5 \textwidth}{!}{%
$
loss_{\{q_i,ind_i\}} = -\log\left(\frac{\exp\left(\cos(q_i,ind_i)\right)}{\sum\limits_{j=1}^{M}\left(\exp(\cos(q_i,ind_j))\right)}\right),
$
}
\end{equation}

where $q_i$ is the embedding of $query_i$, $ind_i$ is the embedding of $index_i$, \begin{math}\cos(q_i,ind_i)\end{math} is the cosine similarity \begin{math}\frac{q_i^T ind_i}{\|q_i\|\|ind_i\|}\end{math} and M is the number of tuples. \\
This objective helps the model to learn the semantic representations by pulling the embeddings of the same tuple as close as possible and pushing the embeddings of different tuples as far as possible \cite{Hadsell}.

Additionally, the contrastive objective is well aligned with alignment and uniformity proposed by \cite{Wang}, who uses them to measure the quality of semantic representations. The alignment calculates the distance of semantically similar tuples. Meanwhile, the uniformity measures how well the semantic representations are uniformly distributed. Eq.\ref{eq1} can guarantee that the distance of semantically similar tuples is close, and the distribution of semantically dissimilar tuples is scattered on the hypersphere \cite{Gao}.

\textbf{Regularization methods} can effectively improve the performance of neural networks \cite{Srivastava}. Since Transformer-based models have demonstrated the outstanding performance in the learning of semantic representation of sentences \cite{Liu,Kaitao}, the traditional simple and effective regularization method Dropout has also been implemented in it.
Attention dropout regularizes the attention weights in Transformers to prevent different contextual feature vectors from adapting to each other. DropHead drops the entire attention head in a certain proportion during training to prevent the multi-head attention model from being dominated by a small number of attention heads\cite{Wangchunshu}. They have shown their effectiveness in translation and text-classification tasks. In this paper, we apply them in semantic search tasks and compare their results with our method.

\section{Regularized Contrastive Learning}

\subsection{Task description}

We begin by formally defining the task of semantic search. Suppose there are M tuples in the dataset: \(<Query_1,index_1>\),\(<Query_2,Index_2>\), \(\cdots\),\(<Query_M,\\
Index_M>\), the task is then to find the top N best indexes according to the query.

In semantic search tasks, we generally encode queries and indexes into sentence representations, and then calculate their cosine similarity scores for comparison or ranking \cite{Rahutomo}. The wildly used objective is contrastive objective such as cross-entropy shown in Eq.\ref{eq1}. 

We use a pre-trained language model as the retrieval model because of its strong capability to learn semantic representations \cite{Liu}. Our method can improve the capability of learning semantic representations from 2 aspects: data augmentation and regulators built with augmented data.

\subsection{Data Augmentation}

Popular methods to augment textual data usually consist in cropping sentences according to some proportions, replacing some words by synonyms or removing some words. With all these methods, it is difficult to guarantee whether the edited sentence is semantically similar to the original sentence. Our method does not directly augment the number of sentences but rather focus on augmenting the number of semantic representations of sentence: the embeddings.

We first train several entropy models based on Eq.\ref{eq2}.
\vspace{-0.5em}
\begin{equation} \label{eq2}
\resizebox{0.8 \textwidth}{!}{%
$
J_i = -\log\left(\frac{\exp(\cos(q_i,ind_i))}{\sum\limits_{j=1}^{M}(\exp(\cos(q_i,ind_j)))}\right) - \phi * \sum\limits_{j=1}^{M}\left(score_{train_{ij}} * \log(score_{train_{ij}})\right)
$
}
\end{equation}

Where \begin{math}i \neq j, score_{train_{ij}} = \frac{\exp(\cos(q_i,ind_j))}{\sum\limits_{k=1}^{M}(\exp(\cos(q_i,ind_k)))}\end{math} is the semantic similarity score of the trainable embeddings, $q_i$ is the semantic embedding of $query_i$, $ind_j$ is the semantic embedding of $index_j$, \begin{math}\phi\end{math} is the weight.

Eq.\ref{eq2} has 2 terms. The first term is the cross-entropy of Eq.\ref{eq1} on the given dataset. This term assumes that $query_i$ is only semantically similar to $index_i$ of the same tuple, and indiscriminately assumes that it is not semantically similar to all $index_j$ of other tuples. Datasets usually do not have information about the exact relationship between the $query_i$ and the $index_j$ of other tuples. The second term is the entropy, which can change the semantic similarity between $query_i$ and all $index_j$ of other tuples. Different \begin{math}\phi\end{math} determines how much the entropy term changes the similarity between $query_i$ and $index_j$ computed by the cross-entropy term. For example, if we choose a \begin{math}\phi > 0\end{math}, when we minimize the Eq. \ref{eq2}, we will minimize the entropy term, which means we enhance the certainty of semantic similarity of the tuple of $<$query, index$>$, and a \begin{math}\phi < 0\end{math} will make the semantic similarity of tuples of $<$query, index$>$ much more uncertain. We can train several entropy models with different \begin{math}\phi\end{math} and then use each of them to generate different semantically similar embeddings for every sentence, ultimately increasing the number of sentences semantic representations.

\subsection{Contrastive Regulator}

Once we have augmented the embeddings for every sentence through several entropy models, we take them into a new contrastive objective as regulators shown in Eq.\ref{eq3} below.
\vspace{-0.5em}
\begin{equation} \label{eq3}
\resizebox{1.0 \textwidth}{!}{%
$
J_i = -\log\left(\frac{\exp(\cos(q_i,ind_j))}{\sum\limits_{k=1}^{M}(\exp(\cos(q_i,ind_k)))}\right) - \mathop{\underline{\sum\limits_{n=1}^{N}\log\left(\frac{\exp(\cos(q_i,Au_{q_i}^n))}{\sum\limits_{k=1}^{M}(\exp(\cos(q_i,Au_{q_k}^n)))}\right)}}\limits_{Regulator} - \mathop{\underline{\sum\limits_{n=1}^{N}\log\left(\frac{\exp(\cos(ind_i,Au_{ind_i}^n))}{\sum\limits_{k=1}^{M}(\exp(\cos(ind_i,Au_{ind_k}^n)))}\right)}}\limits_{Regulator}%
$
}
\end{equation}
Where $Au^n_{q_i}$ is the augmented semantic similar embedding of $query_i$ from entropy model n, $Au^n_{ind_i}$ is the augmented semantic similar embedding of $index_i$ from entropy model n, N is the number of entropy models.

\subsection{Anisotropy problem}
According to \cite{Wang}, when the number of pairs from different tuples approaches infinity, the asymptotics of Eq.\ref{eq3} can be derived as 
\begin{equation} \label{eq4}
\resizebox{1.0 \textwidth}{!}{%
$
\begin{aligned}
&-E_{(q_i,ind_i)\sim\pi_{pos}}\left[\mathbf{q_i}^\top \mathbf{ind_i}\right]+E_{q_i\sim\pi_{data}}\left[\log E_{ind_j\sim\pi_{data}}\left[e^{\mathbf{q_i}^\top \mathbf{ind_j}}\right]\right]+\\
&\sum\limits_{n=1}^{N}\left(-E_{(q_i,Au_{q_i}^n)\sim\pi_{pos}^q}\left[\mathbf{q_i}^\top \mathbf{Au_{q_i}^n}\right]+E_{q_i\sim\pi_{data}^q}\left[\log E_{Au_{q_j}^n\sim\pi_{data}^q}\left[e^{\mathbf{q_i}^\top \mathbf{Au_{q_j}^n}}\right]\right]\right)+\\
&\sum\limits_{n=1}^{N}\left(-E_{(ind_i,Au_{ind_i}^n)\sim\pi_{pos}^{ind}}\left[\mathbf{ind_i}^\top \mathbf{Au_{ind_i}^n}\right]+E_{ind_i\sim\pi_{data}^{ind}}\left[\log E_{Au_{ind_j}^n\sim\pi_{data}^{ind}}\left[e^{\mathbf{ind_i}^\top \mathbf{Au_{ind_j}^n}}\right]\right]\right)\\
&=-E_{(q_i,ind_i)\sim\pi_{pos}}\left[\mathbf{q_i}^\top \mathbf{ind_i}\right]-\sum\limits_{n=1}^{N}E_{(q_i,Au_{q_i}^n)\sim\pi_{pos}^q}\left[\mathbf{q_i}^\top \mathbf{Au_{q_i}^n}\right]-\sum\limits_{n=1}^{N}E_{(ind_i,Au_{ind_i}^n)\sim\pi_{pos}^{ind}}\left[\mathbf{ind_i}^\top \mathbf{Au_{ind_i}^n}\right]+ \\
&E_{q_i\sim\pi_{data}}\left[\log E_{ind_j\sim\pi_{data}}\left[e^{\mathbf{q_i}^\top \mathbf{ind_j}}\right]\right]+\sum\limits_{n=1}^{N}E_{q_i\sim\pi_{data}^q}\left[\log E_{Au_{q_j}^n\sim\pi_{data}^q}\left[e^{\mathbf{q_i}^\top \mathbf{Au_{q_j}^n}}\right]\right]+\\
&\sum\limits_{n=1}^{N}E_{ind_i\sim\pi_{data}^{ind}}\left[\log E_{Au_{ind_j}^n\sim\pi_{data}^{ind}}\left[e^{\mathbf{ind_i}^\top \mathbf{Au_{ind_j}^n}}\right]\right]
\end{aligned}%
$
}
\end{equation}
Where $\pi_{pos}, \pi_{pos}^q$ and $\pi_{pos}^{ind}$ are uniform distributions of pairs of sentences from the same tuple, $\pi_{data}, \pi_{data}^{q}$ and $\pi_{data}^{ind}$ are uniform distribution of dataset, $\mathbf{q_i^\top\mathbf{ind_i}} = \frac{q_i\top ind_i}{\|q_i\|\|ind_i\|}, \mathbf{q_i^\top\mathbf{{Au_q}_i^n}} = \frac{q_i\top {Au_q}_i^n}{\|q_i\|\|{Au_q}_i^n\|}$ and $\mathbf{ind_i^\top\mathbf{{Au_{ind}}_i^n}} = \frac{ind_i\top {Au_{ind}}_i^n}{\|ind_i\|\|{Au_{ind}}_i^n\|}$.\\

The first three terms of Eq.\ref{eq4} make the sentences from the same tuples have similar semantics, while the last three terms make the sentences from the different tuples have dissimilar semantics and can be derived with Jensen's inequality as
\begin{equation} \label{eq5}
\resizebox{1.0 \textwidth}{!}{%
$
\begin{aligned}
&E_{q_i\sim\pi_{data}}\left[\log E_{ind_j\sim\pi_{data}}\left[e^{\mathbf{q_i}^\top \mathbf{ind_j}}\right]\right]+\sum\limits_{n=1}^{N}E_{q_i\sim\pi_{data}^q}\left[\log E_{Au_{q_j}^n\sim\pi_{data}^q}\left[e^{\mathbf{q_i}^\top \mathbf{Au_{q_j}^n}}\right]\right]+\\
&\sum\limits_{n=1}^{N}E_{ind_i\sim\pi_{data}^{ind}}\left[\log E_{Au_{ind_j}^n\sim\pi_{data}^{ind}}\left[e^{\mathbf{ind_i}^\top \mathbf{Au_{ind_j}^n}}\right]\right]\\
&=\frac{1}{m}\sum\limits_{i=1}^{m}\log\left(\frac{1}{m}\sum\limits_{j=1}^{m}e^{\mathbf{q_i}^\top \mathbf{ind_j}}\right)+\sum\limits_{n=1}^{N}\frac{1}{m}\sum\limits_{i=1}^{m}\log\left(\frac{1}{m}\sum\limits_{j=1}^{m}e^{\mathbf{q_i}^\top \mathbf{{Au_q}_j^n}}\right)+\sum\limits_{n=1}^{N}\frac{1}{m}\sum\limits_{i=1}^{m}\log\left(\frac{1}{m}\sum\limits_{j=1}^{m}e^{\mathbf{q_i}^\top \mathbf{{Au_{ind}}_j^n}}\right)\\
\end{aligned}%
$
}
\end{equation}

\vspace{-2em}
\begin{equation} \label{eq6}
\resizebox{0.9 \textwidth}{!}{%
$
\ge \frac{1}{m^2}\sum\limits_{i=1}^{m}\sum\limits_{j=1}^{m}\mathbf{q_i}^\top \mathbf{ind_j}+\sum\limits_{n=1}^{N}\frac{1}{m^2}\sum\limits_{i=1}^{m}\sum\limits_{j=1}^{m}\mathbf{q_i}^\top \mathbf{{Au_q}_j^n}+\sum\limits_{n=1}^{N}\frac{1}{m^2}\sum\limits_{i=1}^{m}\sum\limits_{j=1}^{m}\mathbf{ind_i}^\top \mathbf{{Au_{ind}}_j^n} 
$
}
\end{equation}

Where $m$ is a finit number of samples.

According to \cite{Ethayarajh,Li}, only a few elements of a language model's learned embedding have large values, thus causing anisotropy problem. 
When we minimize Eq.\ref{eq5}, we are actually minimizing the upper bound of Eq.\ref{eq6}, which leads to minimizing $\mathbf{q}^\top \mathbf{ind}, \mathbf{q}^\top \mathbf{Au_q}$ and $\mathbf{ind}^\top \mathbf{Au_{ind}}$. Since they are all almost positive according to \cite{Gao}, their minimization reduces the large value of the embedding, which flattens the embedding and alleviates the anisotropy problem.


\section{Experiments}

We first evaluate the ability of our method to learn sentence representations by conducting experiments on 7 semantic textual similarity (STS) tasks following the work of \cite{Conneau} and then on 2 challenging FAQ datasets: COUGH\footnote{https://github.
com/sunlab-osu/covid-faq} and FAQIR\footnote{https://takelab.fer.hr/data/faqir/}. In the following sections, we first introduce the details of the datasets in section \ref{Datasets}, then the experiments settings in section \ref{Training Details}, and finally the results of our experiments in section \ref{Results}.

\subsection{Datasets} \label{Datasets}

The 7 semantic textual similarity tasks are: STS 2012-2016 tasks \cite{Agirre,Agirre2013,Agirre_2014,Agirre_2015,Agirre_2016}, the STS benchmark \cite{Cer} and the SICK-Relatedness dataset \cite{Marelli}. For each pair of sentences, these datasets provide a semantic similarity score from 0 to 5. We adopt the Spearman’s rank correlation between the cosine similarity and use the entailment pairs as positive instances, and the contradiction pairs as the hard negative instances.

FAQIR contains 4133 FAQ-pairs for training and test. Most of their answers are very long, which poses a challenge to the retrieval model. 

\subsection{Training Details} \label{Training Details}

We start from the checkpoint of pre-trained model of $SRoBERTA_{base}$. Then we separately fine-tune the entropy models and the final model with different contrastive objectives. The main process is shown in the algorithm
\ref{alg:algorithm1}.
\begin{algorithm}
\caption{Regularized Contrastive Learning Workflow}
\begin{algorithmic}[1]
\State \textbf{Fine-tune the entropy model}
\For{\begin{math}\phi = -value1, -value2, \ldots, +value1, +value2, \ldots\end{math}}
  \State \parbox[t]{\dimexpr\textwidth-\leftmargin-\labelsep-\labelwidth}{Fine-tune the entropy model with the Eq.\ref{eq2} as the objective, denoted as $Model_{entorpy}$\strut}
  \State \parbox[t]{\dimexpr\textwidth-\leftmargin-\labelsep-\labelwidth}{ Calculate semantically similar embeddings of each query and index through $Model_{entropy}$, denoted as $Au^n_{q_i}$ and $Au^n_{ind_i}$\strut}
\EndFor
\State \textbf{Fine-tune the final contrastive learning model}
  \State Fine-tune the final model, with the Eq.\ref{eq3} as the objective, denoted as $Model_{RCL}$
\end{algorithmic}
\label{alg:algorithm1}
\vspace{-0.5em}
\end{algorithm}
For the semantic texture similarity tasks, we train the model on both MNLI and SNLI datasets and calculate the Spearman correlation on the development dataset of STS for evaluation. For the FAQ retrieval task, we evaluate the model by calculating the accuracy of TOP1, TOP3, TOP5 and MAP \cite{Karan} with 5-fold method. More training details can be found in Appendix A.

\subsection{Results} \label{Results}

In table \ref{tab:t2}, we present the results of adopting DropAttention and DropHead on 7 semantic textual similarity tasks.

In table \ref{tab:t3}, we present the results of comparing the method of Attention Dropout, DropHead and RCL on 7 semantic textual similarity tasks.

In table \ref{tab:t4}, we present the results of $SRoberta_{base}$, and the results of the implementation of DropHead, Attention Dropout and RCL on COUGH and FAQIR, which show that RCL has a better performance on all the metrics of TOP1, TOP3, TOP5 and MAP.

\subsection{Ablation Study}

We investigate the impact of the number of regulators. All the reported results in this section are the average of Spearman’s correlation on the 7 text semantic similarity tasks.

We use different $\phi$ to train several different entropy models with Eq. \ref{eq2} as the objective. Each entropy model can generate one semantically similar embedding for every sentence by which we can construct 2 regulators with Eq. \ref{eq3}. If we train more entropy models, we will have more different semantically similar embeddings for every sentence. Consequently, the model can learn the embeddings from more regulators using Eq. \ref{eq3} as the objective. The Table \ref{tab:t5} shows how the number of regulators affect the accuracy of the model.

\begin{center}
\begin{table}[h!]
    \begin{subtable}[h]{\textwidth}
        \centering
        \scriptsize
        \begin{tabular}{ |m{2cm}|m{1.1cm}|m{1.1cm}|m{1.1cm}|m{1.1cm}|m{1.2cm}|m{1.1cm}|m{1.1cm}|m{1.2cm}| } 
            \hline
            model & sts12 & sts13 & sts14 & sts15 & sts16 & stsb & sickr & Average \\
            \hline
            $SRoberta_{base}$ & 74.46\% & 84.80\% & 79.98\% & 85.24\% & 81.89\% & 84.84\% & 77.85\% & 81.29\% \\ 
            \hline
            Attention dropout (p=0.1) & \underline{74.53\%} & \underline{85.08\%} & \underline{80.24\%} & \underline{85.60\%} & *82.64\% & \underline{85.43\%} & 77.88\% & 81.63\% \\
            \hline
            Attention dropout (p=0.2) & 73.10\% & 84.49\% & 79.38\% & 85.32\% & 81.99\% & 84.88\% & 78.81\% & 81.14\% \\ 
            \hline
            Attention dropout (p=0.3) & 73.61\% & 84.79\% & 79.44\% & 85.45\% & 81.77\% & 84.81\% & 78.07\% & 81.13\% \\ 
            \hline
        \end{tabular}
        \caption{Results of Attention Dropout method}
        \label{tab:t2a}
    \end{subtable}
    \newline
    \begin{subtable}[h]{\textwidth}
        \centering
        \scriptsize
        \begin{tabular}{ |m{2cm}|m{1.1cm}|m{1.1cm}|m{1.1cm}|m{1.1cm}|m{1.2cm}|m{1.1cm}|m{1.1cm}|m{1.2cm}| } 
            \hline
            model & sts12 & sts13 & sts14 & sts15 & sts16 & stsb & sickr & Average \\
            \hline
            $SRoberta_{base}$ & 74.46\% & 84.80\% & 79.98\% & 85.24\% & 81.89\% & 84.84\% & 77.85\% & 81.29\% \\ 
            \hline
            DropHead (p=0.1) & 73.71\% & 85.04\% & 79.56\% & 85.36\% & 82.10\% & 85.00\% & 77.73\% & 81.21\% \\
            \hline
            DropHead (p=0.2) & 74.02\% & 84.73\% & 79.58\% & 85.43\% & 81.89\% & 84.89\% & 77.86\% & 81.20\% \\ 
            \hline
            DropHead (p=0.3) & 74.18\% & 84.30\% & 79.20\% & 85.18\% & 82.02\% & 84.81\% & 78.21\% & 81.13\% \\ 
            \hline
        \end{tabular}
        \caption{Results of DropHead method}
        \vspace{-2em}
        \label{tab:t2b}
     \end{subtable}
        \caption{Test set results (Spearman’s correlation) of compared models on text semantic similarity task. In each of the tables, the results in the first row are the results of $SRoberta_{base}$ model. The results in the other rows are the results of adopting different ratio of Attention Dropout and DropHead methods on $SRoberta_{base}$ model. The results are reported as the average of 5 random runs. ``\_" and ``$*$" means the statistically significance improvement with $p < 0.1$ and $p < 0.05$. These results show the Attention Dropout with $p = 0.1$ improves the performance. However, DropHead method does not improve the performance.}
    \label{tab:t2}
\end{table}
\end{center}

\begin{center}
\begin{table}[h!]
    \begin{subtable}[h]{\textwidth}
        \centering
        \scriptsize
        \begin{tabular}{ |m{3.5cm}|m{2cm}|m{2cm}|m{2cm}|m{2cm}| } 
            \hline
            model & Acc\_top1 & Acc\_top3 & Acc\_top5 & MAP \\
            \hline
            $SRoberta_{base}$ & 20.74\% & 32.94\% & 39.38\% & 29.90\% \\ 
            \hline
            DropHead (p=0.1) & 20.62\% & 33.22\% & 39.88\% & 29.94\% \\
            \hline
            Attention dropout (p=0.1) & 21.42\% & 34.33\% & 41.05\% & 31.00\% \\ 
            \hline
            RCL(ours) (p=0.3) & *22.01\% & *35.05\% & \underline{41.42\%} & *31.65\% \\ 
            \hline
        \end{tabular}
        \caption{Results on FAQIR}
        \label{tab:t4a}
    \end{subtable}
    \newline
    \begin{subtable}[h]{\textwidth}
        \centering
        \scriptsize
        \begin{tabular}{ |m{3.5cm}|m{2cm}|m{2cm}|m{2cm}|m{2cm}| } 
            \hline
            model & Acc\_top1 & Acc\_top3 & Acc\_top5 & MAP \\
            \hline
            $SRoberta_{base}$ & 5.95\% & 12.62\% & 17.67\% & 12.72\% \\ 
            \hline
            DropHead (p=0.1) & 5.96\% & 12.73\% & 17.75\% & 12.90\% \\
            \hline
            Attention dropout (p=0.1) & 6.14\% & 13.05\% & 17.95\% & 13.05\% \\ 
            \hline
            RCL(ours) & *6.35\% & \underline{13.31\%} & *18.49\% & *13.41\% \\ 
            \hline
        \end{tabular}
        \caption{Results on COUGH}
        \vspace{-2em}
        \label{tab:t4b}
    \end{subtable}
        \caption{Test set results (accuracy) of compared models on FAQIR and COUGH. In each of the 2 tables, the results in the first row are the results of $SRoberta_{base}$. The results in the second row are the best results of the implementation of DropHead method. The results in the third row are the best results of the implementation of Attention Dropout method. The last row are the results of RCL. These results are reported as the average of 5 random runs. ``\_" and ``$*$" means the statistically significance improvement with $p < 0.1$ and $p < 0.05$.}
    \label{tab:t4}
\end{table}
\end{center}

\begin{center}
\begin{table}[h!]
    \begin{subtable}[h]{\textwidth}
        \centering
        \scriptsize
        \begin{tabular}{ |m{2cm}|m{1.1cm}|m{1.1cm}|m{1.1cm}|m{1.1cm}|m{1.2cm}|m{1.1cm}|m{1.1cm}|m{1.2cm}| } 
            \hline
            model & sts12 & sts13 & sts14 & sts15 & sts16 & stsb & sickr & Average \\
            \hline
            $SRoberta_{base}$ & 74.46\% & 84.80\% & 79.98\% & 85.24\% & 81.89\% & 84.84\% & 77.85\% & 81.29\% \\ 
            \hline
            DropHead (p=0.1) & 73.71\% & 85.04\% & 79.56\% & 85.36\% & 82.10\% & 85.00\% & 77.73\% & 81.21\% \\
            \hline
            Attention dropout (p=0.1) & 74.53\% & 85.08\% & 80.24\% & 85.60\% & *82.64\% & 85.43\% & 77.88\% & 81.63\% \\
            \hline
            RCL(ours) & *76.15\% & *85.84\% & *80.89\% & *86.58\% & *83.36\% & 84.85\% & *79.14\% & 82.40\% \\ 
            \hline
        \end{tabular}
    \end{subtable}
     \caption{Test set results (Spearman’s correlation) of compared models on 7 text semantic similarity tasks. The results in the first row are the results of $SRoberta_{base}$. The results in the second row are the best results of DropHead method. The results in the third row are the best results of Attention Dropout method. The results in the last row are the results of RCL. These results are reported as the average of 5 random runs. ``$*$" means the statistically significance improvement with $p < 0.05$. These results show RCL improves the performance of 6 out of 7 text semantic similarity tasks.}
    \label{tab:t3}
    \vspace{-1em}
\end{table}
\end{center}

\begin{center}
\begin{table}[h!]
    \begin{subtable}[h]{\textwidth}
        \centering
        \scriptsize
        \begin{tabular}{ |m{1.2cm}|m{1.7cm}|m{1.7cm}|m{1.7cm}|m{1.7cm}|m{1.7cm}|m{1.7cm}| }
            \hline
            & \makecell{Number of \\
              Regulator: \\ 
              0 (No $\phi$)} 
            & \makecell{Number of \\ 
              Regulator: \\ 
              2 ($\phi$ = 0.01)} 
            & \makecell{Number of \\ 
              Regulator: \\ 
              4 ($\phi$ = 0.01, \\ 
              0.02)}
            & \makecell{Number of \\ 
              Regulator: \\ 
              6 ($\phi$ = 0.01, \\ 
              0.02, \\ 
              0.03)}
            & \makecell{Number of \\
              Regulator: \\ 
              8 ($\phi$ = 0.01, \\ 
              0.02, \\ 
              0.03, \\ 
              0.04)}
            & \makecell{Number of \\ 
              Regulator: \\ 
              10 ($\phi$ = 0.01, \\ 
              0.02, \\ 
              0.03, \\ 
              0.04, \\
              0.05)} \\
            \hline
            Average (Spearman Correlation) & 81.29\% & *81.47\% & *81.98\% & *82.29\% & *82.40\% & 82.42\% \\ 
            \hline
        \end{tabular}
    \end{subtable}
      \caption{Impact of number of regulators. ``$*$" means the statistically significance improvement with $p < 0.05$ compared to the model precedent.}
    \label{tab:t5}
    \vspace{-1em}
\end{table}
\end{center}

\begin{center}
\begin{table}[h!]
    \begin{subtable}[h]{\textwidth}
        \centering
        \scriptsize
        \begin{tabular}{ |m{2cm}|m{1.1cm}|m{1.1cm}|m{1.1cm}|m{1.1cm}|m{1.2cm}|m{1.1cm}|m{1.1cm}|m{1.2cm}| } 
            \hline
            model & sts12 & sts13 & sts14 & sts15 & sts16 & stsb & sickr & Average \\
            \hline
            $Model_{entropy}$ $(\phi=0.01)$ & 75.97\% & 83.49\% & 78.66\% & 84.55\% & 82.35\% & 83.49\% & 78.13\% & 80.95\% \\ 
            \hline
            $Model_{entropy}$ $(\phi=0.02)$ & 74.31\% & 82.77\% & 78.90\% & 84.35\% & 81.55\% & 83.49\% & 76.13\% & 80.22\% \\
            \hline
            $Model_{entropy}$ $(\phi=0.03)$ & 73.30\% & 83.18\% & 79.37\% & 84.28\% & 81.09\% & 83.55\% & 75.47\% & 80.03\% \\
            \hline
            $Model_{entropy}$ $(\phi=0.04)$ & 73.31\% & 82.98\% & 79.27\% & 84.13\% & 81.21\% & 83.26\% & 74.93\% & 79.87\% \\
            \hline
            RCL(ours) & *76.15\% & *85.84\% & *80.89\% & *86.58\% & *83.36\% & 84.85\% & *79.14\% & 82.40\% \\ 
            \hline
        \end{tabular}
    \end{subtable}
      \caption{Test set results (Spearman’s correlation) of compared models on 7 text semantic similarity tasks. The results from the first row to the third row are the results of entropy models. The results in the last row are the results of our method. These results are reported as the average of 5 random runs.  ``$*$" means the statistically significance improvement with $p < 0.05$. These results show RCL has better performance than all the entropy models.}
    \label{tab:t6}
    \vspace{-2em}
\end{table}
\end{center}

From table \ref{tab:t5}, we can see that 8 regulators (4 entropy models) can significantly improve the performance of our model. However, if we continue to increase the number of regulators, for instance up to 10 regulators, we the performance stops improving.

In table \ref{tab:t6}, we present the results of our RCL method and the results of all the entropy models used to build the regulators. The results show that none of the entropy models is better than RCL model. But the regulators built from these entropy models help to enrich the expressive capability of learned embeddings and then alleviate the anisotropic problem.
\section{Conclusion}
In this work we explore a new regularization method for Transformer-based models: Regularized Contrastive Learning. Our method adopts the contrastive objective and semantically similar representation of sentences to build regulators, which help to overcome overfitting and alleviate the anisotropy problem of sentences embeddings. The results of our experiments on 7 semantic textual similarity tasks and on 2 challenging FAQ datasets demonstrate that our method outperforms widely used regularization methods, and prove that our method is more effective in learning a superior embedding of a sentence.


\appendix
\section{APPENDIX}
\input{appendix}

%% file: appendix.tex
\subsection{A Training Details}
The Base model is $SRoberta_{base}$ \cite{Reimers}. We download their weights from SentenceTransformers. $SRoberta_{base}$ is trained on MNLI and SNLI by using the entailment pairs as positive instances and the contradiction pairs as hard negative instances.

\subsubsection{For STS tasks}
We carry out the grid search of batch size $\in$[32,64] and $\phi\in[[0.01],[0.01,0.02],[0.01,0.02,0.03],[0.01,0.02,0.03,0.04],[0.01,0.02,0.03,0.04,$ \\
$0.05]]$. We firstly fine-tune $SRoberta_{base}$ with different $\phi$ on MNLI and SNLI with Eq. \ref{eq2}, by using the entailment pairs as positive instances and the contradiction pairs as hard negative instances. The purpose of this step is to train several entropy models who can generate different semantic similar embedding for every sentence. Since the pre-trained model has already trained on these datasets, the training process will converge quickly. We use the early stopping method to quickly stop the training process if the loss doesn’t decrease within 3 steps. Then we continue to fine-tune the pre-trained $SRoberta_{base}$, by taking the augmented embeddings and the positive instances and entailment instances of MNLI and SNLI into the contrastive objectives with the Eq. \ref{eq3}. We train this model for 1 epoch, evaluate it every 10\% of samples on the development set of STS-B by Spearman-Correlation. When the batch size is 64 and $\phi=[0.01,0.0,0.03,0.04]$, our model achieves the best accuracy.
\subsubsection{For FAQ datasets}
We implement RCL on the same pre-trained model as above. And apply the same grid-search as above for batch size and $\phi$.
We firstly fine-tune an initial model based on the pre-trained $SRoberta_{base}$ for several epochs with Eq. \ref{eq1} and save the model who has the highest MAP value on the development dataset. Based on this initial model, we continue to fine-tune the entropy model with different $\phi$ with Eq. \ref{eq2}. Finally, we fine-tune the final retrieval model based on the pre-trained $SRoberta_{base}$ with the augmented embeddings for several epochs, saving the model who has the highest MAP value on the development dataset. When batch size is 64 and $\phi=[0.01,0.02,0.03,0.04]$, the retrieval model achieves the best accuracy.